\documentclass[conference]{IEEEtran}
\IEEEoverridecommandlockouts
\usepackage{cite}
\usepackage{amsmath,amssymb,amsfonts}
\usepackage{algorithmic}
\usepackage{graphicx}
\usepackage{textcomp}
\usepackage{xcolor}
\usepackage{todonotes}
\usepackage{verbatim}
\usepackage{url} 

\def\BibTeX{{\rm B\kern-.05em{\sc i\kern-.025em b}\kern-.08em
    T\kern-.1667em\lower.7ex\hbox{E}\kern-.125emX}}
\begin{document}

\title{An Automatic Ground Collision Avoidance System with Reinforcement Learning\\
}

\makeatletter
\newcommand{\linebreakand}{%
  \end{@IEEEauthorhalign}
  \hfill\mbox{}\par
  \mbox{}\hfill\begin{@IEEEauthorhalign}
}
\makeatother

\author{\IEEEauthorblockN{Seyyid Osman Sevgili}
\IEEEauthorblockA{
\textit{Turkish Aerospace}\\
Istanbul, Turkey \\
seyyidosman.sevgili@tai.com.tr}
\and
\IEEEauthorblockN{Atahan Cilan}
\IEEEauthorblockA{
\textit{Turkish Aerospace}\\
Istanbul, Turkey \\
atahan.cilan@tai.com.tr}
\and
\IEEEauthorblockN{Mahir Demir}
\IEEEauthorblockA{
\textit{Turkish Aerospace}\\
Istanbul, Turkey \\
mahir.demir@tai.com.tr}
\and
\IEEEauthorblockN{Ozgun Can Yurutken}
\IEEEauthorblockA{
\textit{Turkish Aerospace}\\
Istanbul, Turkey \\
ozguncan.yurutken@tai.com.tr}
\linebreakand
\IEEEauthorblockN{Umit Can Bekar}
\IEEEauthorblockA{
\textit{Turkish Aerospace}\\
Istanbul, Turkey \\
can.bekar@tai.com.tr}
}

\maketitle

\begin{abstract}
This article evaluates an artificial intelligence (AI)-based Automatic Ground Collision Avoidance System (AGCAS) designed for advanced jet trainers to enhance operational effectiveness. In the continuously evolving field of aerospace engineering, the integration of AI is crucial for advancing operations with improved timing constraints and efficiency. Our study explores the design process of an AI-driven AGCAS, specifically tailored for advanced jet trainers, focusing on addressing the AGCAS problem within a limited observation space. The system utilizes line-of-sight queries on a terrain server to ensure precise and efficient collision avoidance. This approach aims to significantly improve the safety and operational capabilities of advanced jet trainers.
\end{abstract}


\section{Introduction}
This paper introduces an Advanced Ground Collision Avoidance System (AGCAS) avionics system which employs a reinforcement learning (RL) model, specifically a customized soft actor-critic (SAC) algorithm, along with contemporary hyperparameter optimization methods. Additionally, the model integrates a custom convolutional neural network (CNN) that enhances the algorithm's performance by appending an additional state to the RL inputs.

The feature extractor layer of the custom SAC algorithm was modified to train our AGCAS system. Previously an identity layer, it has now been replaced with our custom CNN. The integrated CNN utilizes information from a custom terrain server built with a digital elevation map (DEM) \cite{dted2000}. We have developed a lidar like system which we call pseuodo-lidar using the terrain server. In pseudo-lidar, two distinct metrics, height-on-terrain (HoT) and height-above-terrain (HaT), are combined. Pseudo-lidar consists of an N-by-N array of customizable points that emanate from in front of the aircraft, exhibiting a constant-degree arching motion. These lidar points provide an aerial perspective of the ground, integrating spatial information into the states of the RL algorithm. Additionally, we extend the aircraft's spatial awareness by employing RADALT, which consists of three comparable points aimed at the terrain beneath the aircraft.

The observation space consists of the aforementioned pseudo-lidar data, RADALT and aircraft base states (p, q, r, etc.). Additionally, the action space includes command inputs for the two control surfaces: the aileron and the elevator. Given the aircraft's capability to automatically control these commands, we have focused on these specific control surfaces for applicability.

The primary challenge in any reinforcement learning problem is devising an appropriate reward function. In AGCAS, we address three specific issues: avoiding terrain collisions, maintaining wing-level flight in the absence of imminent collisions, and executing these operations without oscillating control actions. To achieve this, we have developed a sequential reward function, where the importance of rewards or penalties for a given property increases as the property nears its target. This reward system introduces a dependency between maintaining level flight and avoiding collisions, ensuring that control surface oscillations are minimized.

Our final model, validated by domain experts and simulation data, demonstrates that the implementation of AI significantly improves the performance of AGCAS. The results show that AI provides a more generalizable, stable, and reliable control system for the aircraft.

In summary, our research enhances traditional control-based AGCAS in advanced jet trainers through the incorporation of sophisticated RL and CNN methodologies. We achieve this by developing a sophisticated terrain server and a contemporary reinterpretation of RL baseline and aerospace engineering domain-specific reward and punishment systems.

Our work has three main contributions:

\begin{itemize}
  \item Integration of a digital terrain server to facilitate a DEM, enabling us to obtain possible collision conditions.
  \item Extension of the state space representation by adding LiDAR inputs constructed using DEM, resulting in a state space that contains both visual and sequential data.
  \item Implementation of a variant of the original SAC algorithm, specifically the CNN-SAC version, to more accurately capture visual features.
\end{itemize}

The rest of the paper is organized as follows: Section 2 provides a brief history of previous works. Section 3 introduces reinforcement learning. Section 4 explains the methodology. Section 5 presents the simulation results. Section 6 concludes the paper.

\section{Previous Works}
Systems for avoiding collisions are essential to preserving safety when using military aircraft. This section summarizes important research on AGCAS with a focus on its creation, assimilation with current aircraft systems, and advancement using machine learning techniques.

Early work in the field was spearheaded by NASA and the Air Force Research Lab through the Automatic Collision Avoidance Technology (ACAT) project \cite{acat}. This project laid the groundwork for collision avoidance methodologies, introducing two key approaches as AGCAS and Autonomous Air Collision Avoidance Systems (AACAS). This work shows that the AGCAS needs both digital elevation and navigation data to operate.

The design and integration of AGCAS into actual flight systems were demonstrated in an early flight test \cite{agcasint}.This study emphasized the necessity of automation to maintain safety by introducing the idea of autonomous systems that do not require pilot involvement. It emphasized how crucial it is to do away with the necessity of manual pilot intervention, which might be crucial in situations involving a lot of stress.

Lockheed Martin and the Air Force Research Lab further contributed to the development of AGCAS, particularly focusing on integrating the system into existing aircraft with analog flight control computers \cite{agcaspre, agcasanalog}. These studies illustrated the versatility of AGCAS technology by describing the difficulties and solutions associated with retrofitting older aircraft.

A study investigating the application of AGCAS in fighter jets attempted to increase its applicability by emphasizing how aggressive maneuvers and trajectory choices could lower mid-air accidents \cite{agcasfighter}. This study demonstrated AGCAS's usefulness in high-speed, high-risk scenarios, hence confirming its importance in modern military aircraft.

More investigation looked into the modification of AGCAS to accommodate lateral maneuvers in heavier aircraft [6]. This effort expanded the breadth and utility of AGCAS by demonstrating its adaptability to a range of aircraft types and sizes.

In order to introduce a compressed system that made use of a smartphone CPU for digital terrain scanning, AGCAS technology was also investigated for small UAVs \cite{agcasuav}. This study demonstrated the efficacy and adaptability of AGCAS in various scenarios by contrasting its use on smaller UAVs with that of F-16 aircraft.

Through longitudinal field research, the acceptance and level of trust in AGCAS were examined. Pilots responded favorably to the study, and the system was highly trusted \cite{agcasresults}. This study demonstrated the efficacy and dependability of AGCAS by showing that the majority of pilots had no concerns about trust.

When the opinions of novice and experienced pilots on AGCAS were examined, it was found that there were no appreciable variations in terms of trust, although it did draw attention to the possibility of complacency among novice pilots \cite{agcascompare}. This research implies that although most people trust AGCAS, there are some situations in which pilot education and training may be required to reduce hazards.

The use of deep reinforcement learning (DRL) and convolutional neural networks (CNN) in AGCAS was explored [10], the authors used a CNN network in their policy to generate DEM. Their observation and action spaces are relatively same with ours but with slight differences like they also gives the topological data to their observations, that makes the state space is significantly large. Which makes such RL problem complex for optimizing the parameters. So they used a genetic algorithm (GA) for optimization.

\section{Background}
As we have mentioned previously, we are using the RL paradigm to learn aerobatic maneuvers for a desired aircraft. Thus, this section aims to provide a short overview of RL. 

Reinforcement learning is one of the three main learning methods of machine learning, the others being supervised learning and unsupervised learning. Reinforcement learning is explained as learning what to do -how to match the current state to the next action- so that that action maximizes a specific numerical reward. One more important difference is the actions to take are not given to the agent, like most of the other machine learning methods, thus it must learn them via trial and error. \cite{andrew2018reinforcement}. 

In RL, the agent, which is a mapping between observation and action space (generally referred as policy), observes a state $s_{t}$ from its environment at time step t. Then the agent decides on an action $a_{t}$. The action acts upon the environment and the total system shifts into the state $s_{t+1}$. The action will cause the environment to generate reward $R_{t+1}$\cite{arulkumaran2017brief}, which is the main feedback used for learning in RL. Generated data (state, action, reward pairs) is used to improve the current policy.

The below figure depicts a simple working principle of RL:
\begin{figure}[htbp]
\centerline{\includegraphics[width=0.5\textwidth]{figures/rl-algo.png}}
\caption{Inner working of RL.}
\label{fig}
\end{figure}

\section{Methodology}
The methodology section of our paper outlines the iterative development process of our AGCAS project, emphasizing the integration of sophisticated AI techniques. We embarked on a series of experiments to enhance the AGCAS for advanced jet trainers. We used the same reward structure during the development process of AGCAS. 

Our AGCAS model's reward strategy is an evolution of the system we developed in our previous project, PARS, updated to meet the unique demands of avoiding ground collisions and managing altitude \cite{pars2024}. To discourage collisions with terrain, we have set a significant penalty of -250 for each collision. This penalty mirrors the greatest positive reward the system can earn for perfect performance, underlining how crucial it is to prevent collisions. Furthermore, we have set a constant -250 penalty for an excessive amount of negative g-load, which is constrained to -2; this is also the greatest positive reward the system can achieve for perfect performance. As a result, our system combines both dense and sparse rewards, leading to quicker and more effective learning for our agent. 

As explained in the Figure \ref{fig:agcas_rewards}, the system uses a sequential reward process that places avoiding collisions above maintaining level flight. This prioritization kicks in whenever LiDAR detects a potential collision, ensuring that dodging terrain is always the aircraft's top concern. Level flight benefits only come into play when the route is free of obstacles. We've set the LiDAR detection period to 10 seconds, striking a balance between reacting to immediate dangers and strategic foresight.

\begin{figure}[htbp]
\centerline{\includegraphics[width=0.5\textwidth]{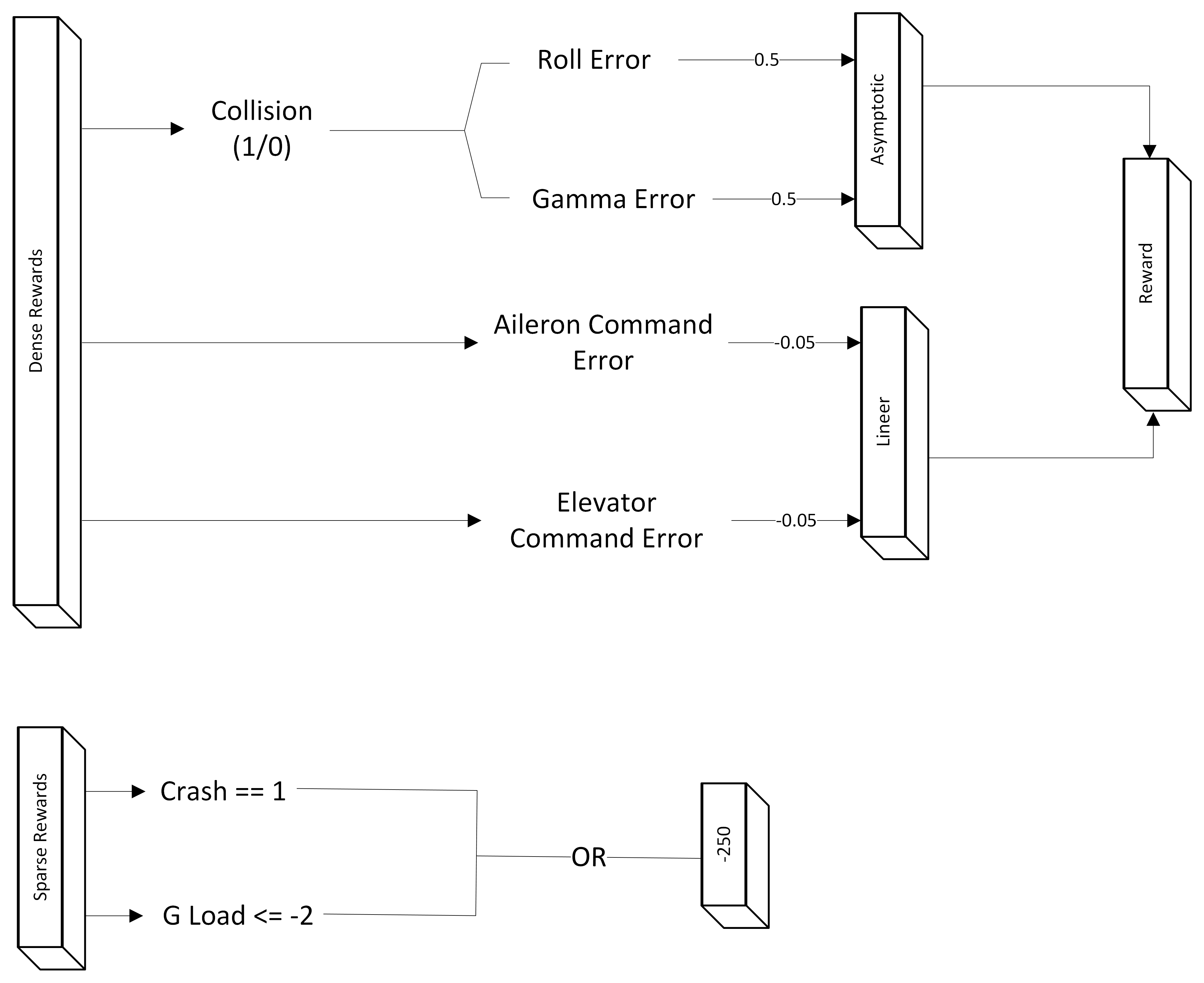}}
\caption{The reward structure.}
\label{fig:agcas_rewards}
\end{figure}

Figure \ref{fig:model_input_output} explains the models inputs and outputs. In this work in addition to our previous study called PARS \cite{pars2024}, we used extra inputs that is KxK LiDAR points and RADALT points to provide visual consciousness to the agent.

\begin{figure}[htbp]
\centerline{\includegraphics[width=0.5\textwidth]{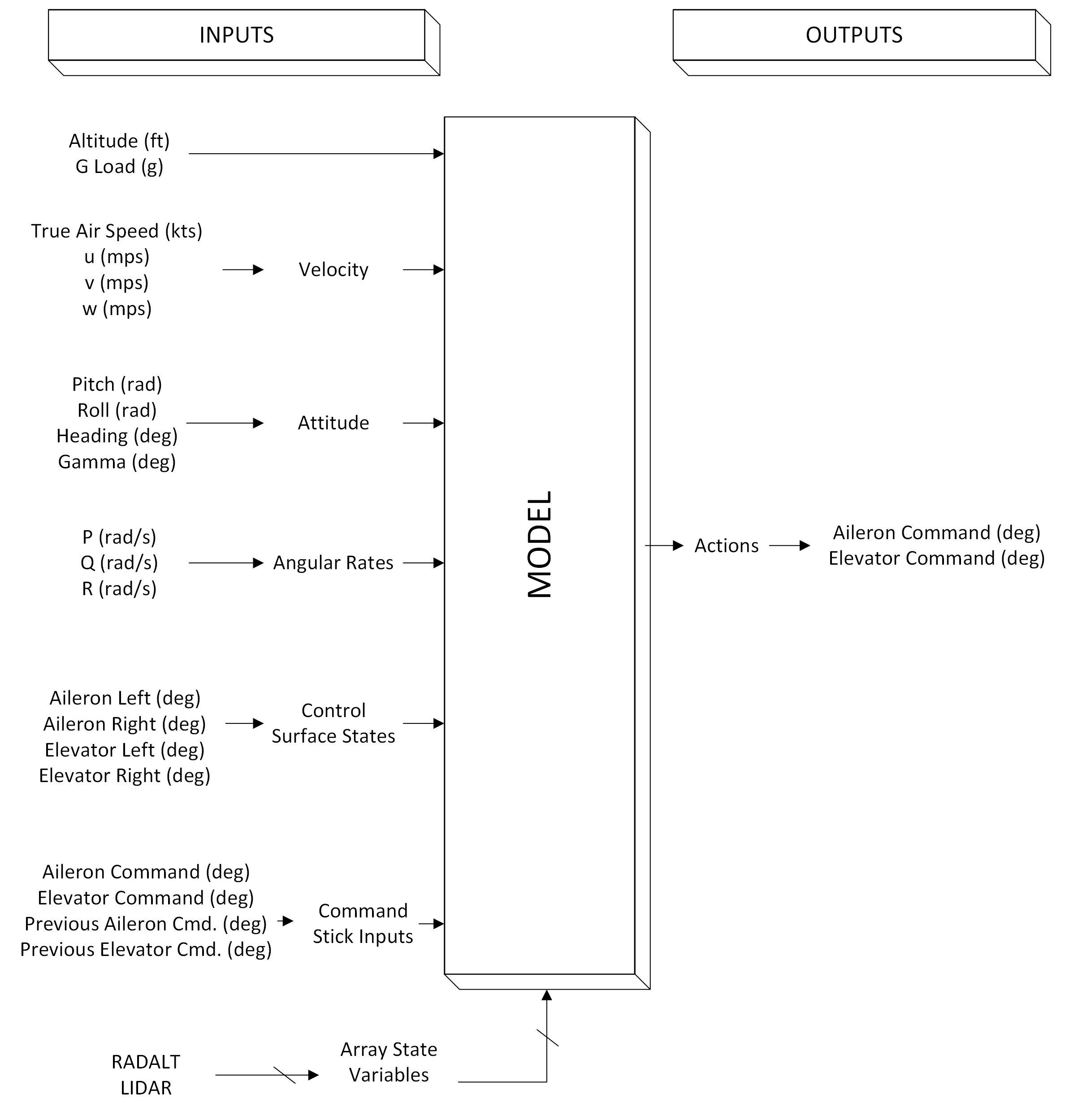}}
\caption{The general model scheme.}
\label{fig:model_input_output}
\end{figure}

\subsection{Initial Condition Generator}

We began by generating appropriate initial conditions for the collision avoidance systems. We developed an algorithm to identify the conditions needed for training. The algorithm receives the coordinates of the search area in latitude and longitude format. Additionally, it takes in the search range for roll and pitch angles, which were both set to 0 and -90 degrees for this study. The algorithm then searches for collisions within a range of 750 meters to 2000 meters, incrementing the roll and pitch angles by 6 degrees in each step separately, while also incrementing the heading by 1 degree at each step. The identified initial conditions are then saved for training purposes. 

\subsection{Single-Point LiDAR and SAC}

Our initial strategy involved integrating a single-point LiDAR input into the state representation used by the Soft Actor-Critic (SAC)\cite{sac-paper} algorithm for training. This approach aimed to leverage LiDAR data for obstacle detection directly in the aircraft's line of sight. However, this method proved insufficient for capturing comprehensive escape strategies, as it offered a very narrow perception of potential obstacles, focusing only on direct threats without considering the broader context necessary for effective avoidance maneuvers.

\subsection{Multi-Point LiDAR and SAC}

Recognizing the limitations of our initial model, we expanded the LiDAR input from a single point to a 3x5 rectangle, thereby broadening the system's perceptual field to include potential escape points. During the development process, LiDAR input generated by setting gaps to 1.5-degree on the vertical sides and 3-degree on the horizontal sides. This adjustment allowed the original SAC algorithm to produce more satisfying results, although it still fell short of our optimal performance criteria which is to perform avoidance maneuver using the most unobstructed area (densest green area on our Figure \ref{fig:lidar_on_terrain}). Further experimentation involved increasing the LiDAR input to a 5x5 square configuration. However, this change did not yield a significant improvement over the previous 3x5 setup, indicating the need for a more sophisticated approach to state representation and feature extraction.

\subsection{Multi-Point LiDAR and CNN-SAC}

The subsequent phase of our methodology focused on developing a custom network structure that could handle both image-like and sequential state representations. We modified the feature extractor layer of the SAC algorithm to incorporate CNN elements, creating a CNN-SAC model. The main motivation for this modification is the preventing the huge expansion of the state space. For instance, for an 16x16 image, we need 256 extra states. This huge expansion will lead the fail of the tasks. Using CNN on the feature extraction phase allows us to represent 256 points with less data. The model explained in Figure \ref{fig:CNN_SAC}. 

This model was designed to process LiDAR inputs configured in a KxK point square, integrating these features with sequential state data to enhance the system's predictive capabilities. Inputs of the CNN layer can be seen clearly on the Figure \ref{fig:lidar}. By examining the Figure \ref{fig:lidar_on_terrain} and Figure \ref{fig:lidar}, we can clearly see that the model are accepting inputs that is created using distances. The nearest point is represented as black and the farthest point is represented as white. In this way we can reduce the 3D data into the 2D to give input to the CNN model. This representation created using our digital terrain server's LOS capability which allows us the get collision data and distance with given range.

Our experiments with various K values led to significant improvements, especially when combined with additional RADALT inputs that provided extra state information from below the aircraft. This comprehensive state representation, processed through our CNN-SAC model, achieved highly satisfying results, confirmed by expert feedback. Expert feedback is one of the most crucial part of this study. With the help of Turkish Aerospace (TA), we can reach the experts easily, which makes the our studies more robust.

\begin{figure}[htbp]
\centerline{\includegraphics[width=0.5\textwidth]{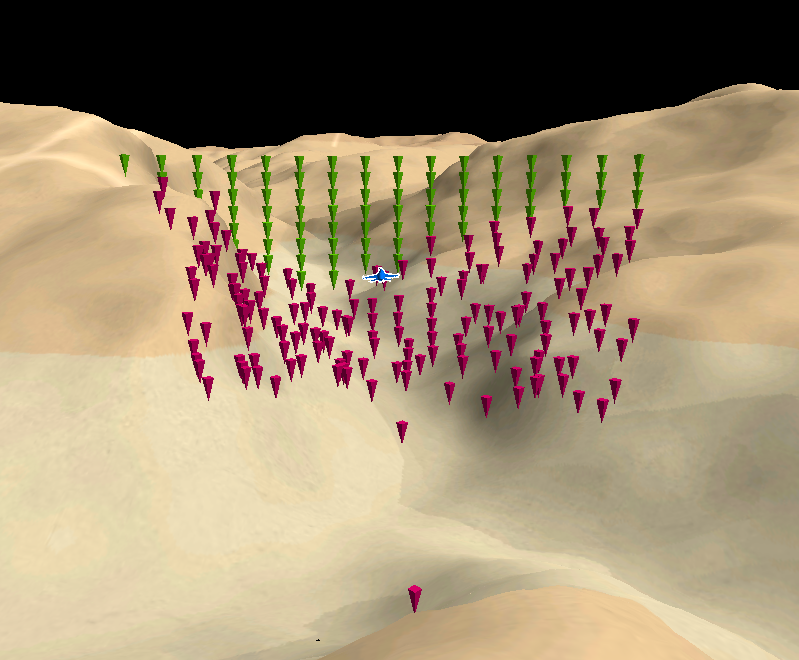}}
\caption{16x16 LiDAR on Terrain.}
\label{fig:lidar_on_terrain}
\end{figure}

\begin{figure}[htbp]
\centerline{\includegraphics[width=0.25\textwidth]{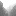}}
\caption{16x16 LiDAR.}
\label{fig:lidar}
\end{figure}


\begin{figure}[htbp]
\centerline{\includegraphics[width=0.5\textwidth]{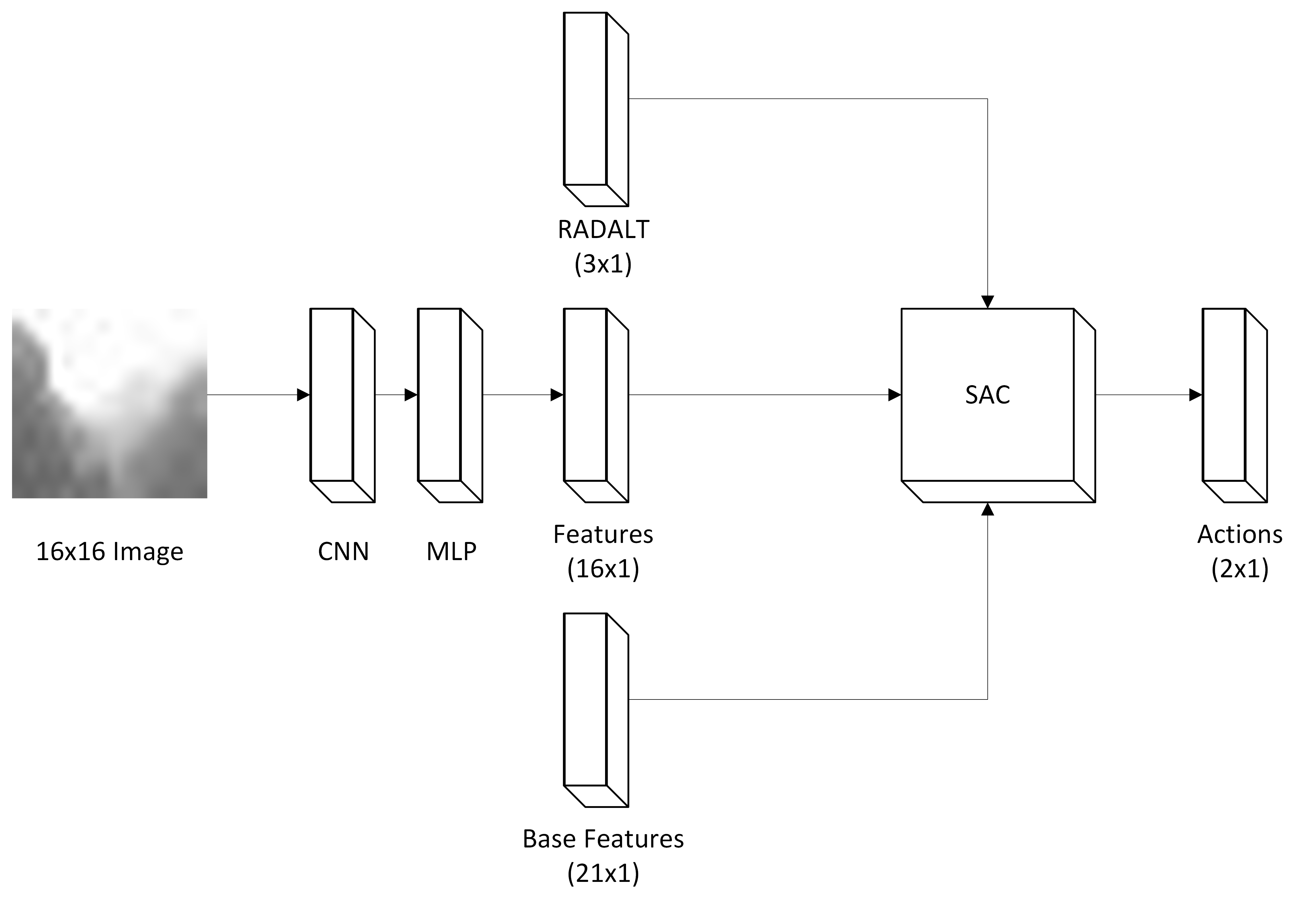}}
\caption{CNN-SAC Structure.}
\label{fig:CNN_SAC}
\end{figure}

\subsection{Exploration of CNN-LSTM-SAC}

In an effort to capture temporal dependencies that could further optimize collision avoidance strategies, we experimented with a CNN-LSTM-SAC model. This approach, however, did not meet our expectations, suggesting potential areas for future research, such as incorporating visual attention mechanisms to improve model performance.

\subsection{Hyperparameter Optimizaiton}
To ensure that our models achieved optimal performance, we conducted comprehensive hyperparameter optimization as the final stage of our study. Hyperparameter tuning can have a crucial impact on performance improvement in machine learning tasks, especially in RL.

For this purpose, we utilized Optuna \cite{akiba2019optuna}, a state-of-the-art open-source hyperparameter optimization framework. Optuna offers an advanced and efficient approach to hyperparameter tuning, incorporating features like automatic search space pruning, which helps efficiently explore the search space.
\section{Results}
\begin{figure*}
\centerline{\includegraphics[width=1\textwidth]{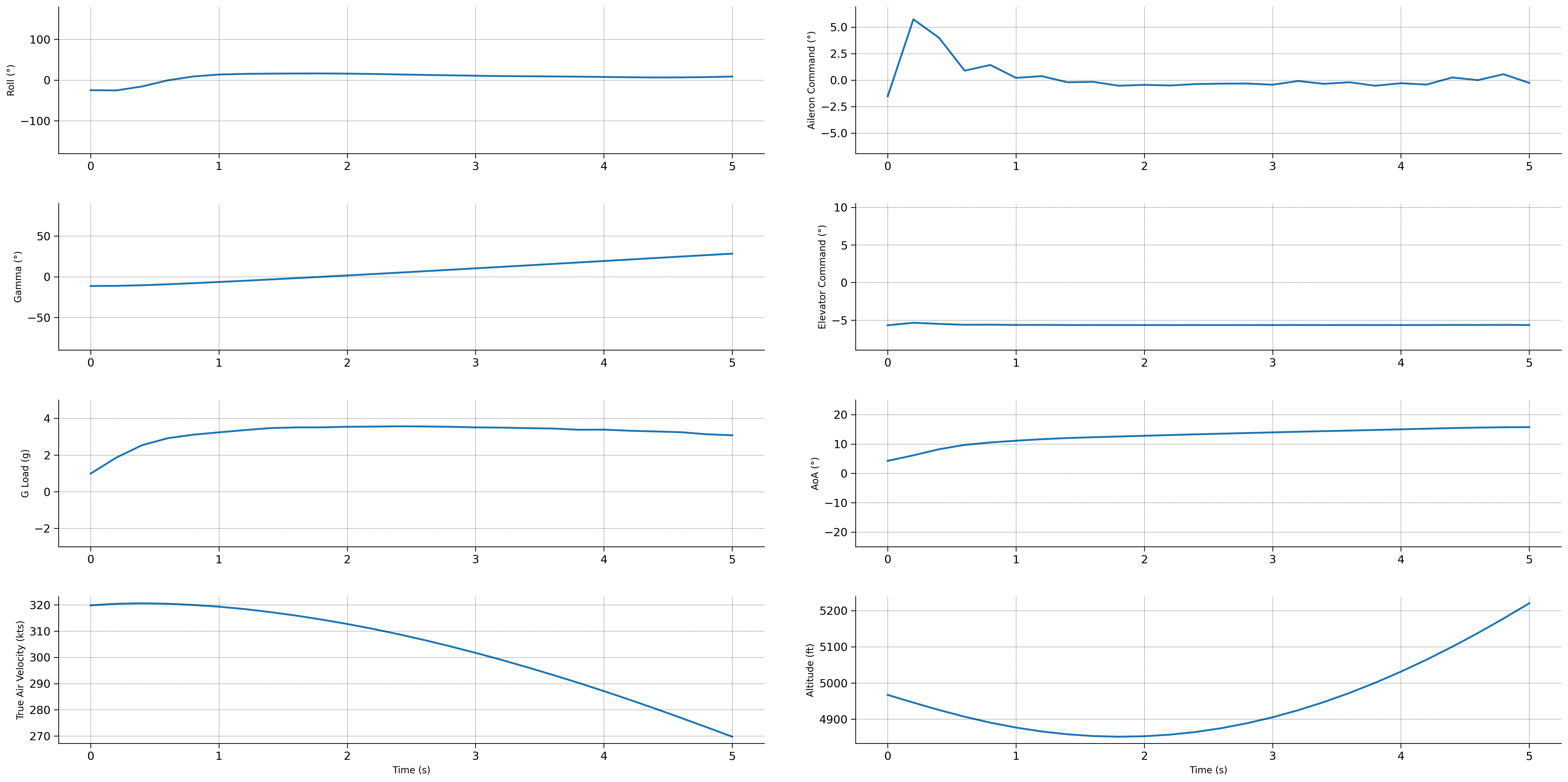}}
\caption{Rescue graphs.}
\label{fig:rescue2}
\end{figure*}

Figure \ref{fig:rescue2} shows the important states and action graphs for the second case, which is shared in  \cite{youtube2021}. Our methodology reflects a rigorous, iterative approach to developing an AI-driven AGCAS. By continuously refining our model through the integration of complex state representations and reinforcement learning (RL), we have improved upon the traditional AGCAS framework. Our findings not only demonstrate the effectiveness of incorporating AI into avionics but also lay the groundwork for future advancements in this critical area of aerospace engineering. 

\section{Conclusion}
In this paper, we proposed a sophisticated CNN-SAC model for the AGCAS. We conducted a series of experiments involving various scenarios, including the use of pure SAC with Single-Point LiDAR, Multi-Point LiDAR, Multi-Point LiDAR with CNN-SAC, and Multi-Point CNN-LSTM-SAC. The results demonstrated that the CNN-SAC model, equipped with a 16x16 LiDAR input, performed the best. The integration of CNN into the SAC feature extraction layer significantly enhanced the agent's ability to interpret the terrain and identify potential escape routes. This improvement facilitated a valid path planning for escape, showcasing the model's effectiveness in navigating complex environments. 
\section*{Acknowledgement}
The project was realized as part of Hurjet advanced jet trainer platform's artificial intelligence concept projects within Turkish Aerospace. The domain experts were employees of the company. The simulation environment and flight dynamics model of the jet were provided from the company.

\bibliographystyle{ieeetr}
\bibliography{bibi}

\end{document}